\documentclass{article}
\usepackage{lettrine}
\usepackage[utf8]{inputenc}
\usepackage[english]{babel}

\usepackage[margin=0.6in]{geometry}

\usepackage{multicol}
\usepackage{authblk}
\usepackage{cite}
\usepackage{graphicx}

\usepackage{float}

\usepackage[ruled]{algorithm2e}

\title{\textbf{Data Segmentation via t-SNE, DBSCAN, and Random Forest}}
\author{Timothy DeLise}
\affil{Université de Montréal}
\date{January 2021}

\begin{document}

\maketitle

\begin{multicols}{2}

\begin{abstract}
This research proposes a data segmentation algorithm which combines t-SNE, DBSCAN, and Random Forest classifier to form an end-to-end pipeline that separates data into natural clusters and produces a characteristic profile of each cluster based on the most important features.  Out-of-sample cluster labels can be inferred, and the technique generalizes well on real data sets. We describe the algorithm and provide case studies using the Iris and MNIST data sets, as well as real social media site data from Instagram. This is a proof of concept and sets the stage for further in-depth theoretical analysis.
\end{abstract}

\begin{figure*}
\begin{center}
    \includegraphics[scale=0.37]{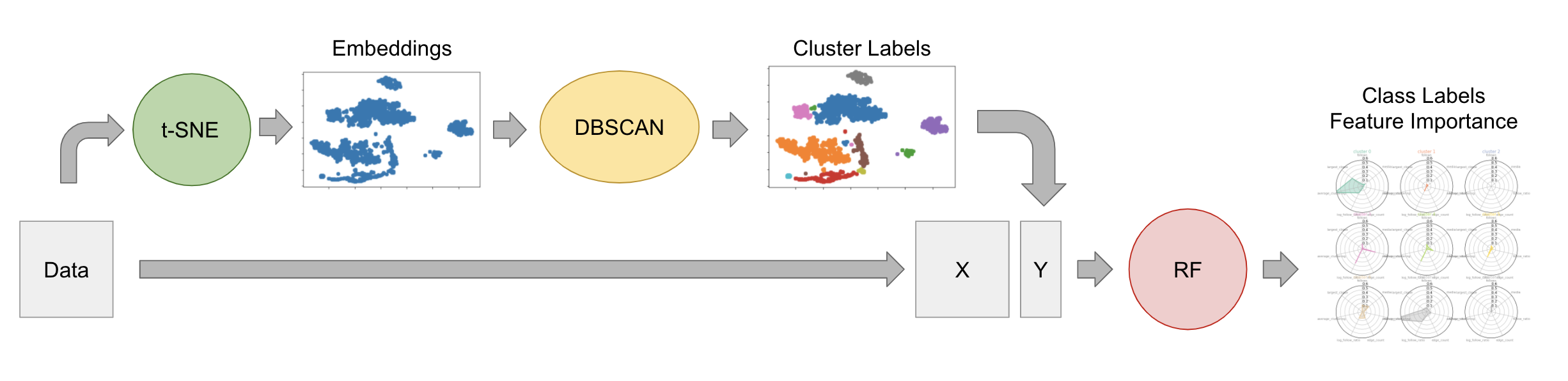}
\end{center}
\caption{This flowchart shows the design of the segmentation analysis algorithm. The data we assume is organized in the usual manner with rows representing individual data points and the columns representing the features of the data. The t-SNE algorithm is applied to the data resulting in a 2-dimensional embedding. Then the DBSCAN algorithm is applied to this embedding, resulting in labeled clusters of the data. Finally these labeled clusters are used as target labels for the Random Forest classifier, using the original (high dimensional) data as input. The Random Forest then has the ability to map data points to cluster labels, and also gives access to feature importance scores. }
\label{flowchart1}
\end{figure*}

\lettrine{\textbf{D}}{ata} segmentation refers to the process of dividing data into clusters and interpreting the characteristics of these clusters, which information can be used for decision making purposes. It is clustering but with an additional requirement to understand the reason behind the clustering and stratification of the data. Data segmentation is widely used in a broad range of fields from social media site marketing\cite{Rajagopal2011} to the analysis of single-cell RNA sequencing\cite{DBLP:journals/corr/abs-1712-09005,Kobak2019}. There are many possible choices of clustering technique as well as possible methods of interpreting the characteristics of each cluster. This research proposes an intuitive, general purpose data segmentation technique which delivers interpretable clusters and tends to generalize well. The algorithm, pictured in Figure \ref{flowchart1}, is comprised of three main steps: t-SNE, DBSCAN, and Random Forest classifier. In the text this process is referred to simply as, \textit{the algorithm}.

T-distributed Stochastic Neighbor Embedding (t-SNE) is the basis of the clustering method. It has been chosen because of its vast popularity in the natural sciences\cite{Kobak2019,Li2017,Platzer2013}, and it is widely regarded as the state of the art for dimension reduction for visualization\cite{mcinnes2020umap}. T-SNE creates a low-dimensional embedding of high-dimensional data with the ability to retain both local and global structure in a single map. It has proven successful for visualizing high-dimensional data\cite{vanDerMaaten2008}. There is strong evidence to support that t-SNE embeddings recover well-separated clusters from the input data\cite{DBLP:journals/corr/LindermanS17}. In practice, high-dimensional data tends to produce distinctly isolated clusters by visual inspection of the low-dimensional output embedding\cite{Kobak2019}. The motivation is to harness the intuitive appeal of the t-SNE embedding. However, t-SNE by itself does not label clusters nor provide information about how and why the clusters appear. Moreover, an aspect of t-SNE that detracts from its ability for inference is that there is no direct map from the input space to the output embedding. 

The algorithm harnesses t-SNE as an intuitive first step to simply visualize the data. Its great appeal is that we can visually inspect a low-dimensional embedding (an image, for example) and manually pick out clusters. In order to automate this process, we use DBSCAN\cite{Ester96adensity-based} to extract clusters directly from this low-dimensional embedding. The reason for choosing DBSCAN, as opposed to other density-based clustering algorithms, is that it has a small number of important parameters and is foundational in the field of density-based clustering. The task of extracting visually-identifiable clusters from data in the plane is something that the DBSCAN algorithm can confidently accomplish.

There is one important parameter of the DBSCAN algorithm, defined in the reference as the \textit{Eps-neighborhood} of a point $p$: $N_{eps}(p)$, that we will simply call $\epsilon$. For specific data, it is possible to select a value for $\epsilon$ that separates dense regions into clusters. Through cross validation, optimal $\epsilon$ values can be discovered. In fact, tuning $\epsilon$ can help recover clusters at different levels of resolution.

The third and final step of the algorithm uses the cluster labels obtained from the DBSCAN algorithm to train a Random Forest Classifier\cite{Breiman2001}. The utility of the Random Forest is two-fold: to infer cluster labels directly from the input data, and to gain access to feature importance scores. Random Forest was chosen as well because of its strong ability to classify data, especially if we have reason to believe the target classes belong to well-separated data points. Moreover, it gives transparency to the question of how the data is separated in the input space via its feature importance scores. This fulfills interpretability requirement of data segmentation.

An important question to investigate is, \textit{how well does the algorithm generalize?} While data segmentation is inherently an unsupervised learning technique, not concerned with ground-truth data labels, there is a way to understand its ability to generalize. Simply put, if we separate our data set into training and test sets, then the algorithm applied to the training set should create the same clusters as the model applied to the entire set. We then compare the cluster labels given to the test set, using the Random Forest from the algorithm trained on the training data, to the cluster labels of the test set given by the algorithm trained on the entire data set. If the algorithm tends to generalize well on experimental data sets. This means that cluster labels of out-of-sample data points can be reliably inferred without retraining the model. It lends evidence that we can trust the feature importance scores of the Random Forest, which describe how and why the clusters are formed.

The structure of this paper is outlined as follows. Section 2 provides the details of the algorithm as well as the technique to assess its generalizability. Section 3 describes empirical examples of the algorithm applied to 3 experimental data sets: the Iris data set, a data set of anonymized Instagram data obtained from previous research\cite{Hassanpour2019}, and the MNIST data set of 70,000 hand-written digits. Sections 4 offers interpretation of the results and motivation for future research.

\section{Methods}
\subsection{The Algorithm}

The algorithm is composed of three main sub-algorithms: t-SNE, DBSCAN, and Random Forest classifier. Figure \ref{flowchart1} gives an overview. We assume that the data is in the usual format, with rows representing individual data points and columns representing the features. T-SNE creates a 2-dimensional embedding of the data. For the next step, the DBSCAN algorithm is applied to the low-dimensional embedding to produce cluster labels for each data point. Finally these cluster labels are used to train a Random Forest classifier via supervised learning. The Random Forest model can thus infer cluster labels directly from the raw input data.

Certain values for $\epsilon$ reveal the clusters which are visually apparent in the t-SNE embedding.  Most values of $\epsilon$ generalize well, although values for $\epsilon$ can be found that generalize extremely well, almost perfectly. In practice we optimize a constant, which is then multiplied by the mean pairwise distance of the t-SNE embedded data points. For more information about $\epsilon$ tuning via cross validation, please refer to section \ref{generalizing_segments}.

The Random Forest admits feature importance scores. These scores allow us to understand which features are most influential in separating the data into clusters.  Combining these scores with \textit{cluster profiles} completes the process of segmenting the data, and hence the algorithm.

\subsection{Cluster Profiles}
\label{cluster_profiles}

\begin{figure}[H]
\begin{center}
    \includegraphics[scale=0.45]{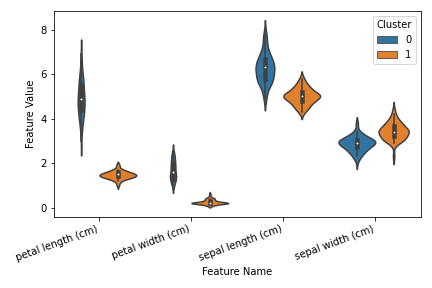}
\end{center}
\caption{The Iris cluster profiles are shown in a \textit{violin plot}, which displays the empirical distribution of the data over each feature, separated by cluster. }
\label{iris violin}
\end{figure}

We define the \textit{cluster profile} to be the distribution of the data points of each feature over that cluster, as in Figure \ref{iris violin}. A simple statistic is the mean value. If our input data has $n$ features, then the cluster profile can be represented is an $n$ dimensional vector of the mean values of each cluster, as shown in Table \ref{arraprofiles1} and Figure \ref{mnistclusterprofiles}. The cluster profile is thus used to characterize the cluster. The feature importance scores of the Random Forest algorithm allow us to focus on the features which matter the most. For example, we quickly understand that petal length is much more important than sepal width, for the purposes of dividing the iris data into clusters (Table \ref{iris feature importance}).

\begin{figure}[H]
\begin{center}
    \includegraphics[scale=0.28]{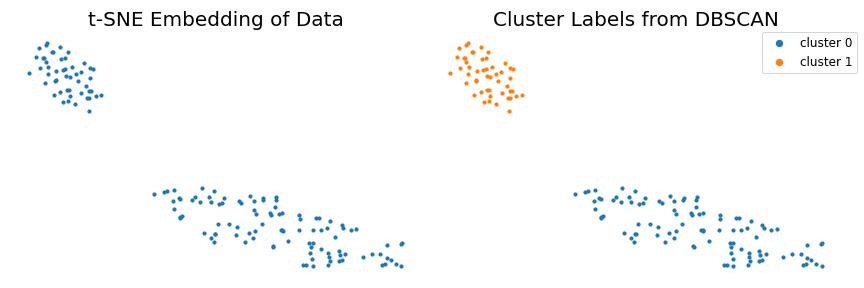}
\end{center}
\caption{The plot on the left is the 2-dimensional embedding that resulted from the t-SNE part of the algorithm. The plot on the right shows the same data points labeled by the cluster labels that were learned using the DBSCAN algorithm applied to the embedding. }
\label{irisplot1}
\end{figure}

\begin{table}[H]
    \centering
    
    \begin{tabular}{c|c}
        Score & Feature Name  \\
         \hline
        0.555780 &  petal length (cm) \\
        0.314322 &   petal width (cm) \\
        0.122197 &  sepal length (cm) \\
        0.007701 &   sepal width (cm)
    \end{tabular}
    \vspace*{2mm}
    \caption{Iris data set feature importance scores calculated by the Random Forest classifier. }
    \label{iris feature importance}
\end{table}

\subsection{Generalizing Segments}
\label{generalizing_segments}

Cluster profiles are developed and we wish for these clusters and their characteristics to generalize to out-of-sample data points. The algorithm gives a way to infer cluster labels of out-of-sample data points using the Random Forest classifier. Here, we describe a technique for assessing the generalizability of the algorithm.

In unsupervised learning scenarios, the data does not contain ground-truth labels, so we take the ground-truth of some particular data point to be the cluster label that is assigned to that data point when the entire data set is run through the algorithm. We then randomly split the whole data set into training (in-sample) and test (out-of-sample) sets in the usual way. In our case we use the 5-fold cross validation technique described in \cite{Breiman1984}. For each fold of data, the algorithm is run on the training set, which returns the cluster labels of the training data and a Random Forest classifier that will map input data to cluster labels. Finally we infer the cluster labels from the test set by applying the Random Forest classifier. Classification metrics are computed using the labels obtained from the test set compared to the \textit{ground-truth} labels that were computed from the entire data set. In Table \ref{generalization performance} the weighted averages of the classification metrics over all 5 folds of the data are displayed. 

Cross validation is used, as well, on each training set to choose the optimal value of the DBSCAN parameter $\epsilon$. This makes the generalization procedure quite computation-intensive since the training set for each fold of the 5-fold cross validation is used to optimize $\epsilon$ by way of 5-fold cross validation. This additional computation time is merited for the purpose of a thorough analysis of the algorithm. In practice the $\epsilon$ parameter can be chosen using less expensive means and validated using cross validation before applying it to the entire data set. We forego the cross validation of the $\epsilon$ parameter in the MNIST experiment for the sake of time savings and additional \textit{resolution} of the clusters. 

The idea of cluster resolution can be illustrated by considering the following thought experiment. A very large value for $\epsilon$ will always produce only one cluster, and this technique will obviously always generalize perfectly. Depending on the data, we may wish to set an lower limit to the number of clusters obtained, thus sacrificing performance for the sake of segmenting the data into more, smaller clusters. This purpose is inherently attained by selecting smaller values for $\epsilon$. By lowering $\epsilon$ we derive more clusters from the data, but this also creates more singleton ( and extremely small ) clusters which detract from the generalizing performance. For the Iris data set we use a lower limit of clusters we require to $2$, and for the Instagram data we set the limit to $5$. In the MNIST data experiment we intentionally set $\epsilon$ small enough to reveal the 10 main clusters of the data, therein creating many small and singleton clusters.

The cluster labels obtained from the analysis of the training data set need not match the labels obtained from the whole data set. The reason is that the cluster label name is chosen somewhat arbitrarily in that we always label the largest cluster as cluster 0, the next largest cluster as cluster 1 and so on. In fact, it is common for the training set to produce a different number of clusters than the whole data set altogether. We have developed a technique to address this by matching the clusters obtained from the training set with those from the entire data set. It is an iterative procedure that matches clusters which have the largest intersection first. Details about this procedure are supplied in Appendix \ref{clustermatchingappendix}.

An alternative technique to compute the out-of-sample classification metrics is to map out-of-sample data points to their embedded location, something that is not possible in the original t-SNE algorithm, however has been implemented in the openTSNE software package\cite{Poliar731877}. We chose not to use this technique in order to focus on the utility of the Random Forest step of the algorithm. However, one should be able to show similar results using the inference mapping of openTSNE.

\subsection{Software}

The software used for the experiments will be made freely available on GitHub. It is a conglomerate of customized code and algorithms with existing software packages. Scikit-learn\cite{scikit-learn} was used for the Random Forest and DBSCAN implementations as well as data scaling and classification metrics. FIt-SNE\cite{Linderman2019} was used for the t-SNE computations as it is fast and has shown success visualizing the MNIST data set.

\begin{table*}
    \centering
    \begin{tabular}{c|c|c|c|c|c}
        Cluster & follows & average shortest path & diameter & clique count & node count  \\
         \hline
        0 &	38.07 &	186.98 &	126.41 &	1.86 &	0.63\\
        1 &	345.61 &	585.47 &	587.62 &	1.90 &	0.68\\
        2 &	0.13 &	0.00 &	0.00 &	0.00 &	0.00 \\
        3 &	113.09 &	264.17 &	180.31 &	0.98 &	0.61 \\
        4 &	122.24 &	328.97 &	260.71 &	1.30 &	0.63 \\
    \end{tabular}
    \vspace*{2mm}

    \caption{Instagram data set cluster profiles. For each of the five clusters we display the mean values of the top five important features over each cluster. }
    \label{arraprofiles1}
\end{table*}

\begin{figure}[H]
\begin{center}
    \includegraphics[scale=0.28]{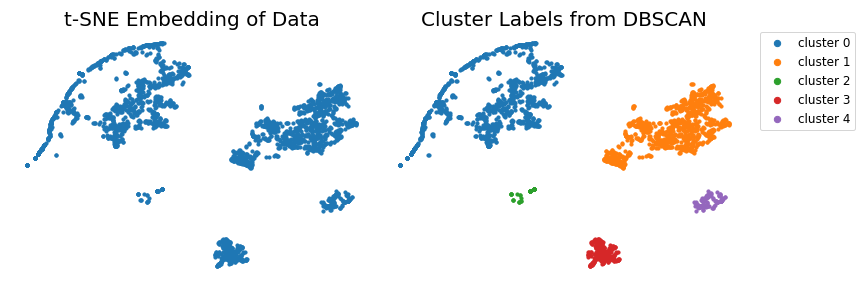}
\end{center}
\caption{The plot on the left shows the 2-dimensional embedding of the Instagram data set that resulted from t-SNE. On the right side is the same embedding with the cluster labels given by the DBSCAN step. }
\label{arraplot1}
\end{figure}

\begin{table}[H]
    \centering
    \begin{tabular}{c|c}
        Score & Feature Name  \\
         \hline
        0.110567 &                follows\\
        0.100257 &  average shortest path\\
        0.091884 &               diameter\\
        0.080912 &           clique count\\
        0.057833 &             node count\\
        0.054657 &            followed by\\
        0.049908 &           follow ratio\\
        0.046561 &      edge connectivity\\
        0.045851 &             edge count\\
        0.044900 &      node connectivity\\
        0.044383 &   average connectivity\\
    \end{tabular}
    \vspace*{2mm}
    \caption{ Instagram data set feature importance table, showing the top ten most important features, ordered by  score. }
    \label{arra_feature_importance}
\end{table}

\section{Empirical Results}
\subsection{Iris Data Set}

The Iris flower data set \cite{Anderson10.2307/2394164,fisher36lda} is a famous, elegant and freely available data set that displays intrinsic clusters. Figures \ref{irisplot1}, \ref{iris violin} and Table \ref{iris feature importance} display the output of the algorithm applied to the Iris flower data set. There are clearly 2 clusters in the data. Table \ref{generalization performance} shows the classification metrics for each generalization experiment.

Referring to Figure \ref{iris violin}, the goal is to understand why the constituents of each cluster have been grouped together. The clusters label is assigned by the number of data points in each cluster. We see that cluster 0 is characterized by longer petal length, petal width, and sepal length than cluster 1 while having shorter sepal width. This simple visualization tool, while by no means exhaustive, already offers substantial insight into the descriptive attributes of each cluster. There is very little overlap between the clusters in the distributions of petal length and width. We understand that petal length and width are more important for inferring these clusters than sepal length and width. This idea matches precisely with the feature importance scores of Table \ref{iris feature importance}. Those familiar with the data set will know that these are measurements from three types of flowers: Iris Setosa, Iris Versicolor, and Iris Verginica. The measurements from Versicolor and Verginica tend to mix while the Setosa is quite separate. It corresponds that the segmentation analysis was able to identify two clusters and not three.


\subsection{Instagram Data}

The analysis of this section follows the same steps as the previous section, the only difference being that we substitute the input data. The data was obtained from a previous study\cite{Hassanpour2019} and is completely anonymized. The features contain simple metrics about Instagram users, such as the number of followers, likes, tags, etc. We also calculated several social network attributes based on the raw data. This data set contains 3,229 data points and 27 features. For more information about the data, please refer to \cite{Hassanpour2019}.

Figure \ref{arraplot1} shows the clustering results from the segmentation analysis. The feature importance scores of Table \ref{arra_feature_importance} combined with the cluster profiles of Table \ref{arraprofiles1} give us the defining characteristics of the clusters. 

Cluster 0 is the largest cluster and cluster 1 is the next largest, corresponding to the blue and orange clusters of Figure \ref{arraplot1} respectively. Cluster 0 is described by data points with less follows, average shortest path, and diameter, and cluster 1 has higher values for these important features. We see cluster 2, which is the third largest cluster, has a mean that is zero or almost zero across the important features. These are seemingly empty accounts. The segmentation analysis makes it easy to understand how the data is stratified.


\begin{figure}[H]
\begin{center}
    \includegraphics[scale=0.28]{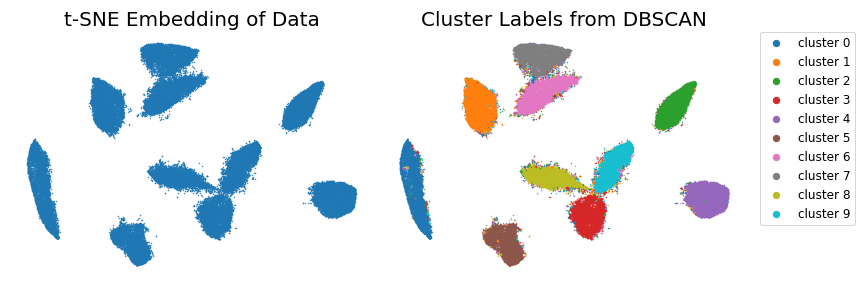}
\end{center}
\caption{MNIST database t-SNE embedding and top ten derived clusters. Notice the many small sporadic clusters that are produced around the edges of the main clusters. }
\label{mnistplot1}
\end{figure}

\begin{figure}[H]
\begin{center}
    \includegraphics[scale=0.28]{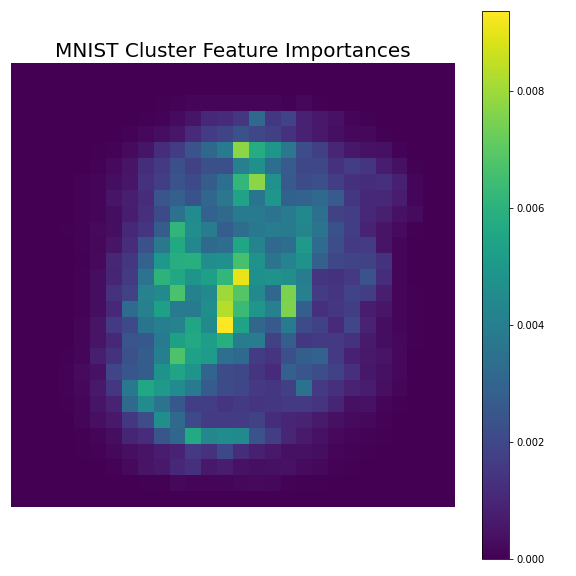}
\end{center}
\caption{A heat map of the feature importance scores learned by the Random Forest step of the algorithm applied to the MNIST data set. In this experiment, features correspond to pixels, so we display the feature scores that correspond to each pixel. We notice that the important features are located toward the center of the image, which is the area of the image where the digits appear. }
\label{mnistfeatureimportances}
\end{figure}

\begin{figure}[H]
\begin{center}
    \includegraphics[scale=0.28]{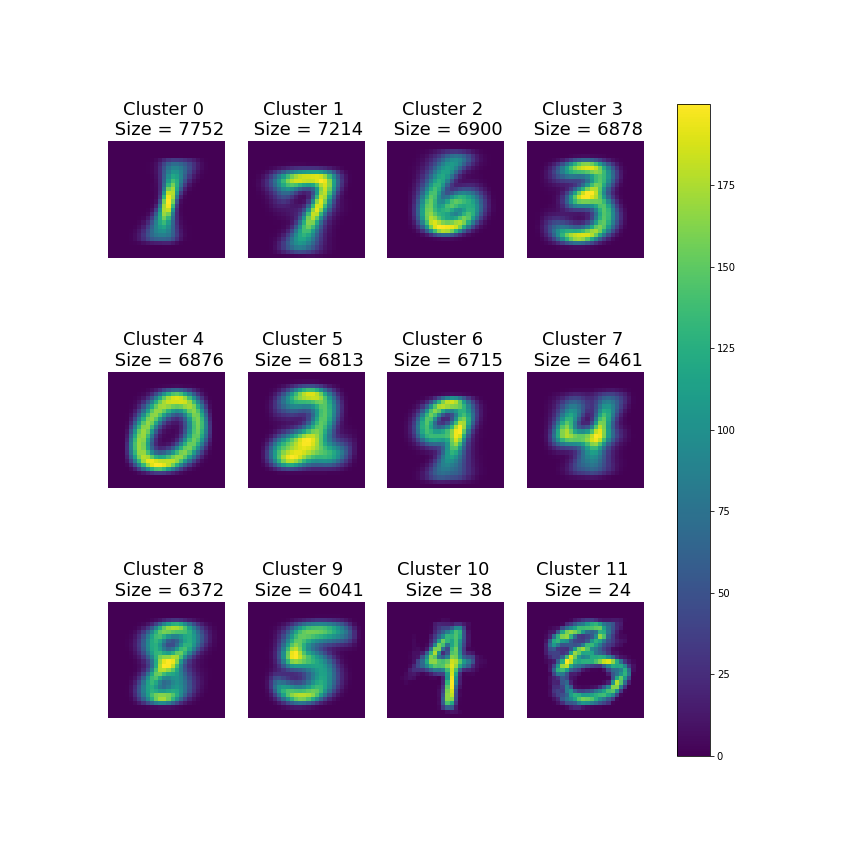}
\end{center}
\caption{The cluster profiles for the top 12 clusters ( by size ) derived from the MNIST data set. The cluster profile is the mean of each feature over the cluster. Just like figure \ref{mnistfeatureimportances}, the features correspond to pixels, so the cluster profiles are displayed as images, where each pixel is the mean of all the respective pixels from each cluster. We notice that the first 10 clusters are all substantially larger than the remaining clusters. This is also apparent from the embedding image of figure \ref{mnistplot1}. Each of the largest ten clusters are representative of each of the ten digits. }
\label{mnistclusterprofiles}
\end{figure}

\subsection{MNIST Case Study}

There are, inevitably, settings for which the default parameters for t-SNE don't quite get us the best embedding. Tuning t-SNE can sometimes produce better visual clusters. The purpose of this section is to illustrate that the algorithm is robust in regards to parameter tuning. Here we address the MNIST data set of 70,000 hand-written images. The embedding produced using the default parameters for t-SNE does not clearly separate ten clusters. However, by using late exaggeration \cite{DBLP:journals/corr/abs-1712-09005}, the authors of \cite{Linderman2019} show that clusters clearly appear in the produced embedding. Although this data set contains 10 distinct data labels corresponding to each of the first ten digits, it has traditionally been difficult for clustering algorithms to clearly identify clusters corresponding to these ten digits.

Even though the embedding on the left side of Figure \ref{mnistplot1} seems to show ten distinct clusters, a few of the clusters are slightly touching in certain regions. $\epsilon$ has been adjusted in order to capture the ten main segments of the data. In doing so a bit of performance was sacrificed in that many very small clusters, often singleton clusters, were identified, which are very difficult to generalize. Nevertheless, we found that the clusters identified in this way still generalize well by weighted average measure.

The feature importance scores highlight the important features that contribute to the separation of the clusters. Since each feature corresponds to a pixel, this conveniently gives an intuitive interpretation where we can visualize the important pixels spatially on a two dimensional image in Figure \ref{mnistfeatureimportances}. The result agrees with our intuition that the middle section of the image should be most important for separating the data into clusters. 

Finally, we visualize the cluster profiles in Figure \ref{mnistclusterprofiles} in image form as well. The ten largest clusters, in fact, correspond to representations of the ten digits. By focusing on the largest clusters and the most important features, we can understand a vast majority of the data.

\subsection{Generalization Performance}

\begin{table}[H]
    \resizebox{\columnwidth}{!}{%
        \centering
        \begin{tabular}{c|c|c|c|c}
            Data Set & Accuracy & Precision & Recall & F1-Score  \\
             \hline
            Iris & 1 & 1 & 1 & 1 \\
            Instagram & 0.943 & 0.996 & 0.943 & 0.967 \\
            MNIST & 0.916 & 0.918 & 0.916 & 0.911
        \end{tabular}
    }
    \vspace*{2mm}
    \caption{The classification metrics for each of the experiments use the weighted average method for calculations. These numbers represent the mean of each weighted score across all five folds of the data. }
    \label{generalization performance}
\end{table}

Table \ref{generalization performance} displays the accuracy, precision, recall and f1-score as a weighted average over all 5 folds of the data. We find for these data sets, the algorithm generalizes well in the sense that out-of-sample data points are most likely going to be classified into the correct cluster by the Random Forest classifier. The good performance emphasizes our belief that the feature importance scores produced by the Random Forest are useful. Moreover, we are confident that the information gained from the algorithm extends to a broader population.

\section{Conclusion and Future Work}

It deserves to be written that this research is superficial in nature and relies on statistical evidence as a proof-of-concept. The author intends to further develop a theoretical understanding. The value of this paper is for the engineer or data science practitioner who needs to get answers from data for which there is little understanding. It relies on the success of t-SNE for visualizing data and adds a layer of interpretability in a practical sense. The additional step is subtle but important for mission-critical applications.

There are some theoretical connections to be made between the t-SNE and DBSCAN step. One of the main results of  \cite{DBLP:journals/corr/LindermanS17} is a theoretical guarantee that all clusters of the input data will be mapped to \textit{balls} in the embedding which can be made arbitrarily small. It is plausible that DBSCAN can successfully identify such \textit{balls} in a low-dimensional space based on the ideas of connectivity and reachability in density-based clustering. The remaining piece is to create a formal argument that if the DBSCAN algorithm has identified a cluster in the embedding space, then this must correspond to a cluster in the input space. This will be a topic of future research.

Random Forest has been a robust supervised learning tool for a long time. If we guarantee that we have labeled actual clusters in the input data, which should follow from the previous paragraph, then we should expect the Random Forest classifier to be able to successfully classify these data points. There should be a way to statistically guarantee that Random Forests can classify disjoint clusters of data. This is another direction of future research. The outline given in these two paragraphs should deliver a more substantial theoretic argument for why this algorithm can dependability be used for data segmentation. 

\appendix
\section{Matching Clusters for Generalization Analysis}
\label{clustermatchingappendix}

This section describes the algorithm used to match clusters between the entire data set and the training data sets that are split during each fold of the generalization analysis. As mentioned in the text, we perform the equivalent of 5-fold cross validation to calculate the average weighted f1-score across all 5-folds of the data. During each fold, we needed a technique to pair the clusters derived from training data with the clusters from the entire data set. This comes down to matching cluster labels, since the labels assigned to each cluster do not necessarily match between runs of the algorithm.

The effect of this matching is really very subtle. Let us do a simple thought experiment by considering the Iris data set, where we saw two main clusters. Something that could happen is that the clusters derived from the entire data set are labeled cluster 0 and cluster 1. The clustering results from the training data during one of the folds of could have derived 2 main clusters as well, however the algorithm could have labeled cluster 0 as cluster 1, and cluster 1 as cluster 0. The matching outlined in this section simply gives us a quick technique to match those labels. 

The technique here is also robust to the situation where the training data derives a different number of clusters than the entire data set. Algorithm \ref{matchingalgo} will match as many clusters as it can, in a largest-first fashion. The optimal algorithm would consider all the permutations of the clusters of the training data compared to the entire data set, aiming to maximize the intersection of all the clusters, however this can require too many computations on large data sets. We sacrifice a bit of performance in terms of f1-score in lieu of considerable time benefits.

\begin{algorithm}[H]
\SetAlgoLined
\KwResult{ BestPerm is list that maps the cluster labels from the AllClusters to TrainClusters. The index position of BestPerm corresponds to the cluster label number of TrainClusters, and the value in that position corresponds to the cluster label of AllClusters. }
TrainClusters is a list of clusters from training data\;
AllClusters is a list of clusters from the entire data set\;
\For { Cluster1 in TrainClusters }{
    BestSum = 0\;
    BestCluster = None\;
    \For { Idx, Cluster0 in AllClusters }{
        ThisSum = Size of Intersection of Cluster1 and Cluster0\;
        \If{ ThisSum is greater than BestSum and Idx not in BestPerm }{
            BestSum = ThisSum\;
            BestCluster = Idx\;
        }
    }
    BestPerm.append(BestCluster)\;
}

\caption{Algorithm for Matching Cluster Labels Between Entire Data Set and Training Data Set}
\label{matchingalgo}
\end{algorithm}



\bibliographystyle{IEEEtran}
\bibliography{mainbib}

\end{multicols}

\end{document}